\documentclass[10pt,twocolumn,letterpaper]{article}

\usepackage{cvpr}
\usepackage{times}
\usepackage{epsfig}
\usepackage{graphicx}
\usepackage{amsmath}
\usepackage{amssymb}
\usepackage{hhline}
\usepackage{multirow}
\usepackage{diagbox}
\usepackage{algorithm}
\usepackage{algorithmicx}
\usepackage[noend]{algpseudocode}
\algdef{SE}[DOWHILE]{Do}{doWhile}{\algorithmicdo}[1]{\algorithmicwhile\ #1}%
\usepackage{tabularx}
\usepackage{comment}
\usepackage{enumerate}
\usepackage{enumitem}
\usepackage{booktabs}

\usepackage[pagebackref=true,breaklinks=true,letterpaper=true,colorlinks,bookmarks=false]{hyperref}
\cvprfinalcopy 

\def\cvprPaperID{2961} 

\ifcvprfinal\fi
\begin{document}

\title{Learning a Weakly-Supervised Video Actor-Action Segmentation Model with a Wise Selection}

\author{Jie Chen \quad Zhiheng Li \quad Jiebo Luo \quad Chenliang Xu\\
Department of Computer Science, University of Rochester\\
{\tt\small \{jiechen, zhiheng.li, jiebo.luo, chenliang.xu\}@rochester.edu}
}

\maketitle
\thispagestyle{empty}


\begin{abstract}
We address weakly-supervised video actor-action segmentation (VAAS), which extends general video object segmentation (VOS) to additionally consider action labels of the actors. The most successful methods on VOS synthesize a pool of pseudo-annotations (PAs) and then refine them iteratively. However, they face challenges as to how to select from a massive amount of PAs high-quality ones, how to set an appropriate stop condition for weakly-supervised training, and how to initialize PAs pertaining to VAAS. To overcome these challenges, we propose a general Weakly-Supervised framework with a Wise Selection of training samples and model evaluation criterion ($\mathcal{WS}^2$). Instead of blindly trusting  quality-inconsistent PAs, $\mathcal{WS}^2$ employs a learning-based selection to select effective PAs and a novel region integrity criterion as a stopping condition for weakly-supervised training. In addition, a 3D-Conv GCAM is devised to adapt to the VAAS task. Extensive experiments show that $\mathcal{WS}^2$ achieves state-of-the-art performance on both weakly-supervised VOS and VAAS tasks and is on par with
the best fully-supervised method on VAAS.
\end{abstract} 

\section{Introduction} \label{sec:intro}

Video actor-action segmentation (VAAS) has recently received significant attention from the community~\cite{can_xu_cvpr15, actor_xu_cvpr16, weakly_yan_cvpr17, joint_kalogeiton_iccv17, learning_zhaofan_transmm18, end_ji_eccv18, actor_kirill_cvpr18}. Extended from general video object segmentation (VOS) which aims to segment out foreground objects, VAAS goes one step further by assigning an action label to the target actor. Spatial information within a single frame may be sufficient to infer the \textit{actors}, but it alone can hardly distinguish the \textit{actions}, \eg, running v.s. walking. VAAS requires spatiotemporal modeling of videos. A few existing works have addressed this problem using supervoxel-based CRF~\cite{actor_xu_cvpr16}, two-stream branch~\cite{joint_kalogeiton_iccv17, end_ji_eccv18}, Conv-LSTM integrated with 2D-/3D-FCN~\cite{learning_zhaofan_transmm18}, 3D convolution involved Mask-RCNN~\cite{end_ji_eccv18}, or under the guidance of a sentence instead of predefined actor-action pairs~\cite{actor_kirill_cvpr18}. Although these fully-supervised models have shown promising results on the Actor-Action Dataset (A2D)~\cite{can_xu_cvpr15}, the scarcity of extensive pixel-level annotation prevents them from being applied to real-world applications.

\begin{figure}
	\begin{center}
		\includegraphics[width=\columnwidth]{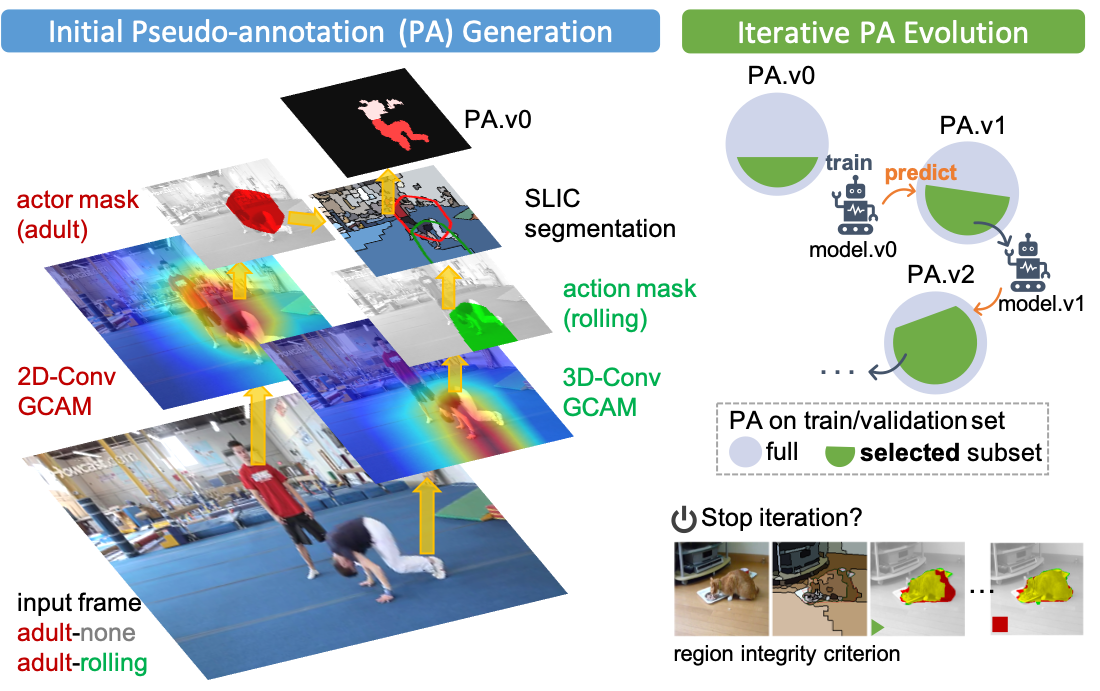}
	\end{center}
	\vspace {-0.2cm}
	\caption{\textbf{Two-stage $\mathcal{WS}^2$ for weakly-supervised VAAS.} Stage-1 (left): Given only a video-level actor-action label, 2D-Conv and 3D-Conv GCAM output actor and action masks (binarized from the actor- and action-guided attention maps). The union of the masks is refined by SLIC~\cite{slic_radhakrishna_pami12}, thus providing a rough location of the target actor doing a specific action for the whole training set. This constructs the initial version of PA (PA.v0). Stage-2 (right): PA evolves through the \textit{select-train-predict} iterative cycles. First, we select a high-quality subset from the latest version of PA to train a segmentation network. The well-trained model is used to predict the next version of PA. When the model's region integrity criterion on the validation set converges, the iteration terminates.}
	\label{fig:meth_framework}
	\vspace{-10pt}
\end{figure}

We approach VAAS in the weakly-supervised setting where we only have access to video-level actor and action tags, such that model generalization is boosted by benefiting from abundant video data without fine-grained annotations. The only existing weakly-supervised method on A2D we are aware of is by Yan \etal~\cite{weakly_yan_cvpr17}. Their method replaces the classifiers in~\cite{actor_xu_cvpr16} with ranking SVMs, but still uses CRF for the actual segmentation, which results in slow inference.

We consider weakly-supervised VOS, a more widely-studied problem. To fill the gap between full- and weak-supervision, a line of works first synthesize a pool of pseudo-annotations (PAs) and then refine them iteratively~\cite{boxsup_dai_iccv15, spftn_dingwen_cvpr17, learning_liang_pami17}. This synthesize-refine scheme is most related to our work but faces the following challenges:

\noindent\textbf{Challenge 1: How to select from a massive amount of PAs high-quality ones?}
In general, PAs are determined by unsupervised object proposals~\cite{key_yong_iccv11, selective_jasper_ijcv13, convolutional_kevis_pami18}, superpixel / supervoxel segmentations~\cite{slic_radhakrishna_pami12, approach_chenglong_tip16}, or saliency~\cite{revisiting_wenguan_cvpr18} inferred from low-level features. Hence, they can hardly handle challenging cases when there is background clutter, objects of multiple categories, or motion blur. The VOS performance is largely limited by the PA quality for models lacking a PA selection mechanism~\cite{weakly_zhou_cvpr18, weakly_tokmakov_eccv16}, or simply relying on hand-crafted filtering rules~\cite{fusionseg_jain_cvpr17, weakly_qizhu_eccv18} that can barely generalize to broader cases. To tackle this challenge, we make a learning-based wise selection among massive PAs rather than blindly trusting the entire PA set. We will show that with only about 25\%-35\% of the full PAs, the selected PAs manage to provide more efficient and effective supervision to the segmentation network that outperforms the full-PA counterpart by 4.46\% mean Intersection over Union ($mIoU$), a relative 22\% improvement, on the test set (see Table \ref{tab:ablation_study}). Note that there is another selection criterion in \cite{multiple_siyang_bmvc17, spftn_dingwen_cvpr17, zigzag_xiaopeng_cvpr18} with a focus on easy/hard samples, whereas ours is good/bad. They are quite different. 

\noindent\textbf{Challenge 2: How to select an appropriate stop condition for weakly-supervised training?}
In supervised training, it is safe to stop training upon the convergence of validation $mIoU$. However, it gets complicated when the obtained validation $mIoU$ is no longer reliable when calculating against the PAs due to the complete absence of the real ground-truth. Fixing the number of training iterations is a simple yet brute solution~\cite{bootstrapping_shen_cvpr18, weakly_tokmakov_eccv16}. Instead, we propose a novel no-reference metric---region integrity criterion ($RIC$)---that does not blindly trust PAs and injects certain boundary constraints in the model evaluation. The convergence of $RIC$ acts as the stop condition in training. Moreover, it turns out that the model with the highest $RIC$ always produces better PAs of the next version than the model with the highest $mIoU$ computed with PAs (see Table \ref{tab:PA_eval_miou}).


\noindent\textbf{Challenge 3: How to initialize PAs when actions are considered along with actors?}
This is a question pertaining to VAAS. Recent works in weakly-supervised image segmentation~\cite{object_wei_cvpr17, bootstrapping_shen_cvpr18, weakly_qizhu_eccv18} and VOS~\cite{weakly_hong_cvpr17} have shown that gradient-weighted class activation mapping (GCAM)~\cite{grad_selvaraju_iccv17} is capable of generating initial PAs from attention maps. However, GCAM is implemented with the network composed of 2D convolutions and trained on object labels; we denote this type of GCAM as 2D-Conv GCAM. Hence, it can only operate on video data frame-by-frame as on images. The spatiotemporal dynamics cannot be captured by 2D-Conv GCAM. Motivated by the success of 3D convolutions~\cite{quo_joao_cvpr17} in action recognition, we extend 2D-Conv GCAM to 3D-Conv GCAM to generate action-guided attention maps that are eventually converted to PAs with action labels. 

In brief, we propose a general \textbf{W}eakly-\textbf{S}upervised framework with a \textbf{W}ise \textbf{S}election of training samples and model evaluation criterion, instead of blindly trusting  quality-inconsistent PAs. We thereby name it $\mathcal{WS}^2$, and Figure~\ref{fig:meth_framework} depicts the framework. In Stage-1, attention maps are generated for each frame and subsequently refined to produce sharp-edged PAs for initializing a segmentation network. In Stage-2, we devise a simple but effective select-train-predict cycle to facilitate PA evolution. Upon a novel region integrity criterion, the performance of video segmentation is enhanced monotonically.

\begin{figure}
	\centering
	\includegraphics[width=1.0\linewidth]{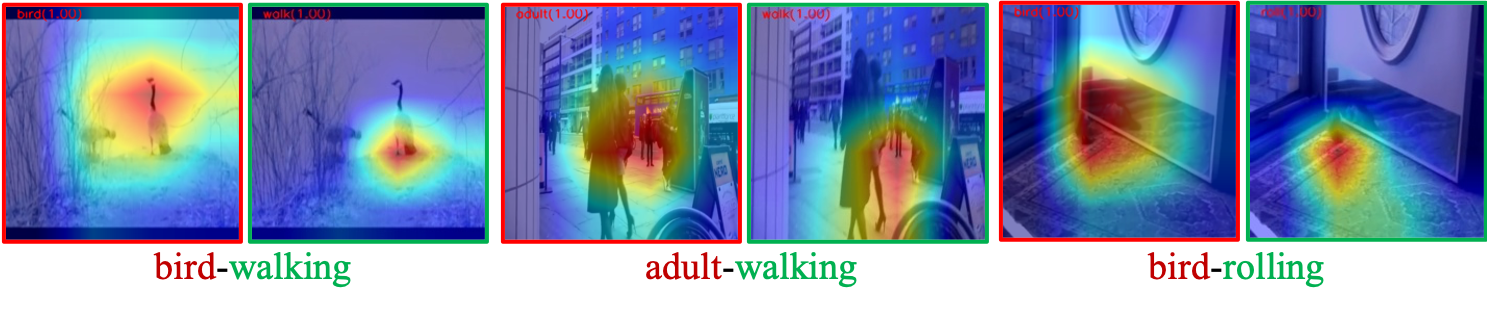}
	\caption{\textbf{Attention maps guided by actor and action.}}
	\label{fig:1-1_intro}
	\vspace{-12pt}
\end{figure}

Customizing the above general-purposed $\mathcal{WS}^2$ to the specific VAAS task is achieved by adding an action-guided attention map obtained by the proposed 3D-Conv GCAM, which has played a complementary role to its 2D counterpart in capturing the motion-discriminate part in the frames. For example, in the first two pairs of Figure~\ref{fig:1-1_intro}, where a bird and an adult are walking, the 3D-Conv GCAM trained on action classification finds the area around legs most discriminative in identifying \textit{walking}, regardless of the appearance difference between adult's and bird's legs, as the motion related to walking always resides on legs.


In summary, our contributions are as follows:
\setlist{nolistsep}
\noindent
\begin{itemize}[noitemsep, leftmargin=*]
	\item We propose an effective two-stage framework $\mathcal{WS}^2$ for weakly-supervised VOS with a novel select-train-predict cycle and a new region integrity criterion to ensure a monotonic increase of segmentation capability.
	\item We customize $\mathcal{WS}^2$ with a 3D-Conv GCAM model to reliably locate the motion-discriminate parts in videos to generate PAs with action labels for the VAAS task.  
	\item Our model achieves state-of-the-art for weakly-supervised VOS on the YouTube-Object dataset \cite{discriminative_kevin_cvpr13, supervoxel_suyog_eccv14}, as well as weakly-supervised VAAS on A2D, which is on par with the best fully-supervised model~\cite{end_ji_eccv18}.
\end{itemize}


\section{Related Work}
In addition to the aforementioned weakly-supervised models of the synthesize-refine scheme for VOS or object detection ~\cite{multiple_peng_cvpr17, w2f_yongqiang_cvpr18}, we also summarize other non-refinement literature on weakly-supervised video object segmentation, as well as action localization.

\noindent
\textbf{VOS.} \quad Motion cue is a good source of knowledge. Using optical flow, Pathak \etal~\cite{learning_pathak_cvpr17} group foreground pixels that move together into a single object, and set it as the segmentation mask to train a model. Similarly, the PAs in \cite{spftn_dingwen_cvpr17} are initialized from segmentation proposals using optical flow and fed into a self-paced fine-tuning network. Tokmakov~\etal~\cite{weakly_tokmakov_eccv16} propose to obtain the foreground appearance from motion segmentation and object appearance from the fully convolutional neural network trained with video labels only. The appearances are then combined to generate the final labels through graph-based inference. However, we try to avoid optical flow in our design due to its inability to handle large motion and brightness constancy constraint. 

The use of non-parametric segmentation approaches is also common. For instance, mid-level super-pixels/voxels are extracted for weakly-supervised semantic segmentation~\cite{weakly_suha_aaai17} and human segmentation~\cite{learning_liang_pami17}. Tang \etal~\cite{normalized_meng_cvpr18} enforce a Normalized Cut (NC) loss in weakly-supervised learning. Since it is accompanied by relatively strong supervision---scribbles, the model has already achieved a $85\%$ full-supervised accuracy even without NC loss.
Similarly, in~\cite{on_meng_eccv18}, shallow regularizers, \ie, relaxations of MRF/CRF potentials, are integrated into the loss. Hence the model could do away with the explicit inference of PAs. 

\noindent
\textbf{Action localization.} \quad
Mettes \etal~\cite{localizing_pascal_bmvc17} introduce five cues, i.e., action proposal, object proposal, person detection, motion, and center bias, for action-aware PA generation and
a correlation metric for automatically selecting and combining them. In contrast, our proposed 3D-Conv GCAM is much simpler by wrapping everything in a unified model.


\section{$\mathcal{WS}^2$ for Weakly-Supervised VOS} \label{sec:vos}

In this section, we illustrate how we design the two-stage $\mathcal{WS}^2$ framework for weakly-supervised video object segmentation. The first stage provides the initial version of pixel-wise supervision on the full training set. The second stage continually improves PAs by iterations of select-train-predict cycles. In each cycle, a portion of more reliable PAs are selected to train a segmentation network, which, in turn, goes through an inference pass to predict a new version of PAs, and a new cycle starts all over again. The whole iteration stops when the highest $RIC$ in each cycle is converged. The overall $\mathcal{WS}^2$approach is shown in Algorithm~\ref{alg: ws^2}.
\begin{algorithm}[t]
    \small
	\caption{$\mathcal{WS}^2$ for weakly-supervised VOS}
	\label{alg: ws^2}
	\begin{algorithmic} [1] 
		\Require weakly-labeled video frames $\{f_i\}$, trained classifier $\Phi$
		\State \textit{\# Stage-1: Initial PA generation}
		\For{$f \in \{f_i\}$}
		\State Generate attention map $S = \textrm{GCAM}(\Phi, f)$
		\State Generate initial mask $M_{\textrm{init}} = \textrm{Otsu}(S)$
		\State Generate refined mask $M_{\textrm{refine}} = \textrm{SLIC\_Refine}(M_{\textrm{init}})$
		\EndFor
		\State $\textrm{PA.v0} = \{M_{\textrm{refine}}\}$
		\State \textit{\# Stage-2: Iterative PA evolution}
		\State Set current version $i = 0$
		\Do
		    \State Select a subset of high-quality PA$_\textrm{select}$ from PA.v$i$
		    \State $RIC^i_\textrm{max} = 0$ \Comment{The maximum $RIC$ achieved at v$i$}
		    \Do
		        \State Train model.v$i$ with PA$_\textrm{select}$
		        \State Evaluate model.v$i$ using $RIC$\Comment{for current epoch}
		        \If{$RIC > RIC^i_\textrm{max}$}
		            \State $RIC^i_\textrm{max} = RIC$
		        \EndIf
		    \doWhile{$RIC$ on the validation set not converge}
		    \State Produce new version PA.v$i$++ by model.v$i$ with $RIC^i_\textrm{max}$ 
		\doWhile{$RIC^i_\textrm{max}$ on the validation set not converge}
		\State \textbf{return} model.v$i$
	\end{algorithmic}
\end{algorithm}
\subsection{Initial Pseudo-Annotation Generation} \label{sec:pseudo_gt}

We first apply 2D-Conv GCAM~\cite{grad_selvaraju_iccv17} to the video frames to locate the most appearance-discriminate regions with a classification network trained on the object labels. Training frames are uniformly sampled over a video. 
The obtained attention map is subsequently converted to the binary mask $M_{\textrm{init}}$ using Otsu threshold~\cite{a_nobuyuki_transsmc79}, which produces the optimal threshold such that the intra-class variance is minimized.

Note that the attention maps calculated from the last convolutional layer are of low-resolution (typically of size 16x smaller than the input size using ResNet-50), the resultant $M_{\textrm{init}}$ is mostly a blob, which can hardly serve as qualified PAs to provide segmentation network with supervision that precisely pinpoints the borders of objects. This issue naturally suggests the use of simple linear iterative clustering (SLIC)~\cite{slic_radhakrishna_pami12}, a fast low-level superpixel algorithm known for its ability to adhere to object boundaries well. We impose the $M_{\textrm{init}}$ on the SLIC segmentation map, thus treating $M_{\textrm{init}}$ as a selector of the superpixels $\{p_{i}\}$. The superpixel selection process is described in Algorithm~\ref{alg: slic_refine}. 
\begin{algorithm}[t]
    \small
	\caption{Mask refinement}
	\label{alg: slic_refine}
	\begin{algorithmic} [1] 
		\Require initial mask $M_{\textrm{init}}$, SLIC  superpixels ${\{p_{i}\}}$, 
		$\alpha$, $\beta$
		\State $P_{\textrm{select}}= \varnothing $
		\For{$p \in \{p_{i}\}$}
		\If {$IoU(p, M_{\textrm{init}}) > \alpha$}
			\If {$R^{ \textrm{area}}_{ p_{i}}= \frac{\textrm{Area}({ p_{i})}}{\textrm{Area}(frame)} < \beta$}
			\State $P_{\textrm{select}}$ add $p$
			\EndIf
		\EndIf
		\EndFor
		\State $M_{\textrm{refine}} = \bigcup P_{\textrm{select}}$
		\State \textbf{return} $M_{\textrm{refine}}$
	\end{algorithmic}
	\vspace {-2pt}
\end{algorithm}

The basic idea is to select superpixel ${p_{i}}$ with sufficient overlap with $M_{\textrm{init}}$ (\textit{line 3}), meanwhile ${p_{i}}$ is not likely to be a background superpixel (\textit{line 4}). Some overly-large foreground objects may be rejected by \textit{line 4}, but there is a tradeoff between high recall and high precision. For PA.v0, we aim to construct a more precise PA for the network to start with. Results in Figure~\ref{fig:test_pred_evol} (l-R) show that our model manages to gradually delineate the entire body of large objects. Finally, the union of the selected superpixels constructs the refined mask $M_{\textrm{refine}}$.

\begin{figure}
	\centering
	\includegraphics[width=1\linewidth]{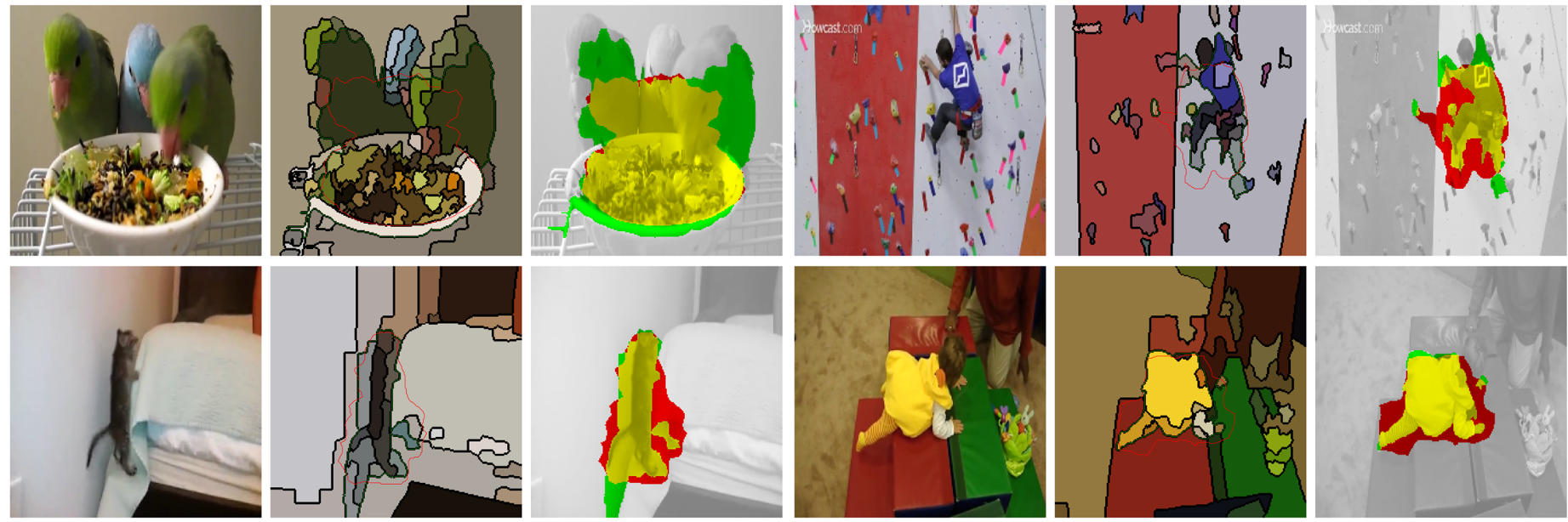}
	\caption{\textbf{Visual results of the mask refinement algorithm.} For each group (left$\rightarrow$right): input frame, SLIC segmentation map, mask refinement results with initial masks being red $\cup$ yellow and refined mask being green $\cup$ yellow.}
	\label{fig:slic_refine}
	\vspace{-0pt}
\end{figure} 

Figure~\ref{fig:slic_refine} shows how the superpixel selection process refines the initial blob-like mask: the false positive part (red) is removed, while the false negative part (green) is successfully retrieved. Such refinement imposes an effective boundary constraint on PA, which is very critical to the dense (pixel-wise accurate) segmentation task.

\subsection{Iterative PA Evolution} \label{sec:iter_train}

The quality of PA.v0 is inconsistent over the full training set, because some challenging cases can hardly be addressed in the initial PAs. To improve the overall quality of PAs, we design a select-train-predict mechanism. First, a subset of PAs is selected to train a segmentation network. Once the network is well-trained, it will make its predictions on the full training set as the new version of PA. The same select-train-predict procedure repeats iteratively until the $RIC$ is converged. 

\begin{figure}
	\centering
	\includegraphics[width=1\linewidth]{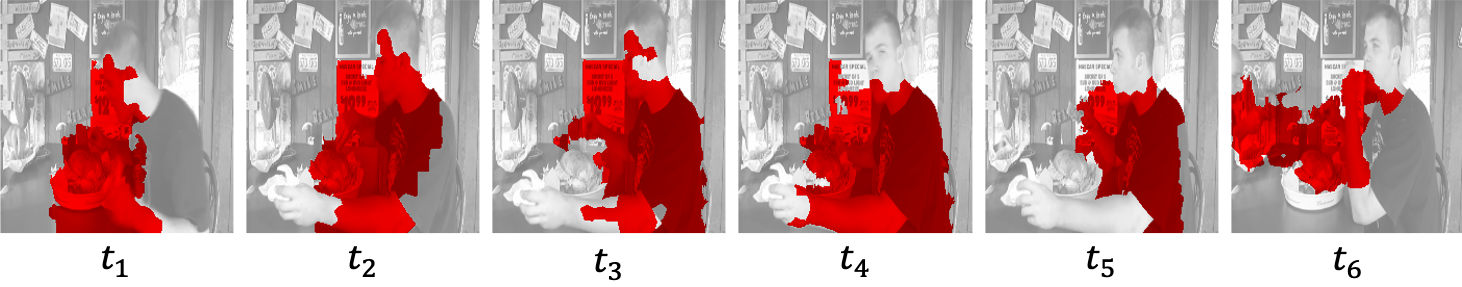}
	\caption{\textbf{Training samples selected from PA.v0 by the relaxed criterion.} PAs among neighboring similar frames provide supplementary information of the object to recover its full body.}
	\label{fig:neighbor_supp}
	\vspace{-12pt}
\end{figure} 

\noindent
{\bf Selection criteria.} \quad The PAs are recognized as of high quality if they either cover the entire object with a sharp boundary (\textit{strict criterion}), or cover the most discriminate part of the object (\textit{relaxed criterion}). Satisfying the relaxed criterion means that a classifier is easy to predict its type if only pseudo-annotated foreground part is visible. This lenience seems to risk taking inferior PAs to training samples as shown in Figure~\ref{fig:neighbor_supp}. However, these samples are still valuable, because they provide abundant training samples with precise localization. And, its inaccuracy can be remedied by the temporal consistency in video data, because by aggregating the information in the adjacent similar frames, the segmentation network could still learn to piece up the full body of the object despite of the noise in annotations.
Taking the video clip in Figure~\ref{fig:neighbor_supp} as an example, the missed arm in frame $t_5$ can be retrieved from the neighboring frame $t_4$.

\begin{figure}
	\centering
	\includegraphics[width=1\linewidth]{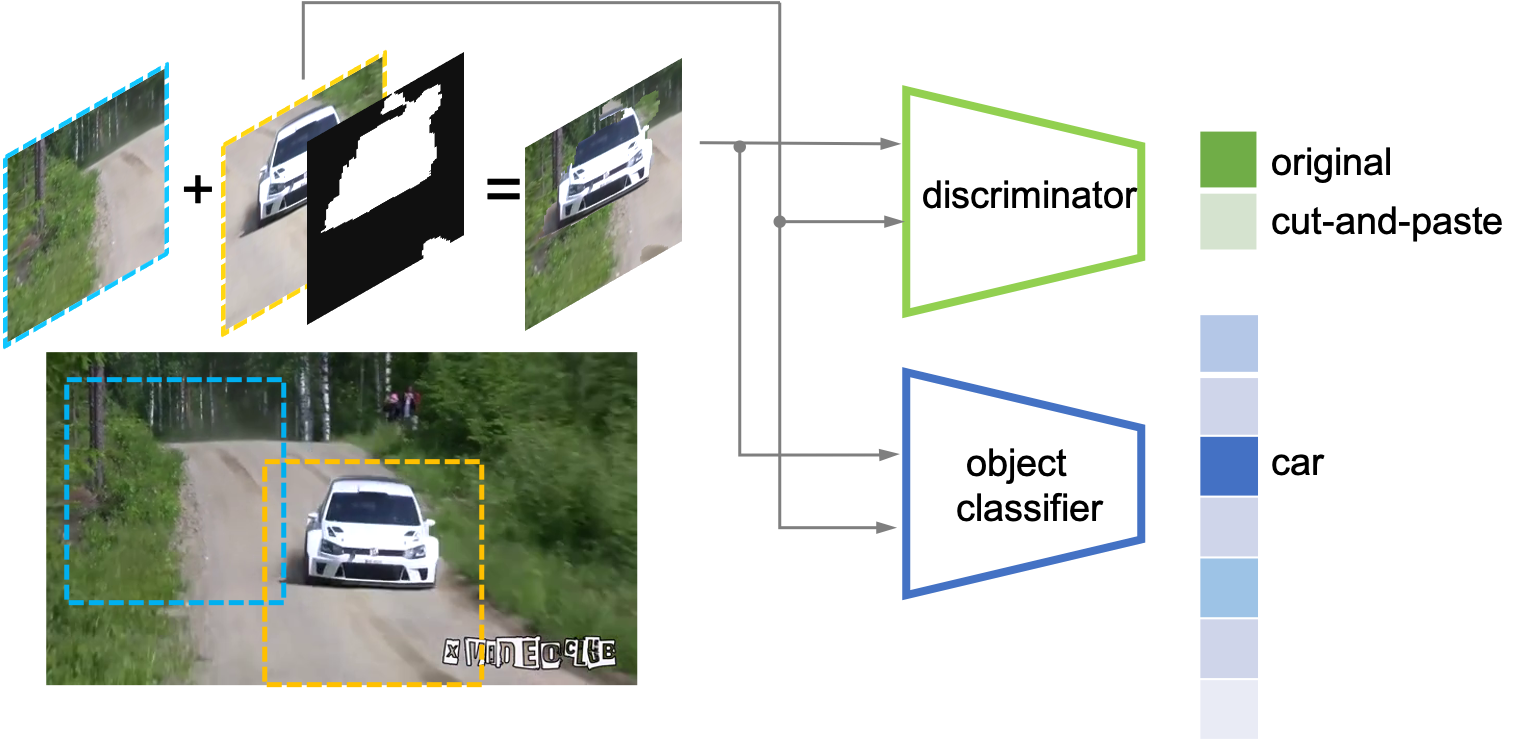}
	\caption{\textbf{Select a subset of high-quality PAs for training.} For the latest version of PA, we can generate a foreground patch (orange-dash) that encloses the object and a corresponding randomly-cropped background patch (blue-dash). Then we cut the object using the PA mask and paste it onto the background patch to construct a cut-and-paste patch. If this patch passes either the test of the binary discriminator or the object classifier, its PAs will be selected to train the segmentation network.}
	\label{fig:cut-and-paste}
	\vspace{-20pt}
\end{figure}

To select the training samples using the above criteria, we employ two networks---a cut-and-paste patch discriminator and an object classifier, as shown in Figure~\ref{fig:cut-and-paste}. Inspired by~\cite{learning_remez_eccv18}, the samples qualified for the strict criterion will cover the whole object with a clear boundary. With such a mask, we can crop out the foreground object and paste it to another background region extracted from the same video, and the cut-and-paste patch still looks real to the binary discriminator. However, the samples matching the relaxed criterion are easily denied by the discriminator, so we add an object classifier to identify them. As long as the mask unveils a certain discriminative part of the object, it will send a strong signal to the object classifier and guide it to recognize its object category.


To prepare the inputs to these two networks, we first sample sets of foreground patches $\{p_\textrm{fg}^i\}$ and background patches $\{p_\textrm{bg}^i\}$ for each video. Foreground patches are squares enclosing the pseudo-annotated objects, and background patches are those not containing any pseudo-annotated objects (Background patches are mostly close to the frame boundary or from frames of scenic shots). Each foreground patch is coupled with a background patch of the same size. Note that they do not necessarily need to come from the same frame but need to be from the same video. It is particularly useful in close shots, where the foreground nearly occupies the full portion of the frame, or when there are multiple objects. In these cases, there is hardly enough space in the same frame for its paired background patch. 

\noindent \textbf{Relation to Remez \etal~\cite{learning_remez_eccv18}.} \quad 
Note that our framework is different from the image-based cut-and-paste model \cite{learning_remez_eccv18} in two ways. First, their weakly-supervised model is under the bounding-box level of supervision, whereas our task is of higher complexity with video-level labels. Second, they employ a GAN framework, where the generator tries to refine the input bounding-box to a tight mask under the guidance of GAN loss, whereas our model is an iterative evolution framework, in which the discriminator plays the role of a selector to pick high-quality PAs for training. There is no generator or adversarial loss involved in our framework, which eases the training process.

\subsection{Region Integrity Criterion (RIC)} \label{sec:ric}

Without the supervision of real ground-truth, it is hard to evaluate the trained model properly. At the end of each select-train-predict cycle, if only mean Intersection over Union ($mIoU$) calculated using PAs is considered to evaluate the model on the validation set:
\begin{equation} \label{eq:miou_pa}
\begin{aligned}
mIoU_\textrm{PA} = mIoU({M_{\textrm{refine}}}, \textrm{PAs})
\enspace,
\end{aligned}
\end{equation}
where $M_{\textrm{refine}}$ is the refined network prediction, it may risk misleading the network to learn the noises in PAs as well. 

In the absence of ground-truth annotation for reference, we thereby introduce a new no-reference metric called Region Integrity Index ($RII$). This metric to-some-extent estimates how much the prediction has recovered the full body of the foreground objects from a low-level perspective. As shown in Figure~\ref{fig:slic_refine}, the initial masks can be refined by SLIC superpixels to fit the boundary of the object. If the difference of the masks before and after the refinement is minor, then it indicates that $M_{\textrm{init}}$ is already fairly precise. Therefore, we define $RII$ in a way that measures how close $M_{\textrm{init}}$ is to its refined version $M_{\textrm{refine}}$:
\begin{equation} \label{eq:rii}
\begin{aligned}
RII = mIoU(M_{\textrm{init}}, {M_{\textrm{refine}}})
\enspace.
\end{aligned}
\end{equation}

The trained models are thus evaluated with region integrity criterion ($RIC$) that combines $mIoU_\textrm{PA}$~\footnote{To distinguish, $mIoU_\textrm{GT}$ is calculated based on the real ground-truth.} with $RII$:
\begin{equation} \label{eq:ric}
\begin{aligned}
RIC = mIoU_\textrm{PA} + \alpha*RII
\enspace,
\end{aligned}
\end{equation}
where $\alpha = 0.5$ in our setting. Such design incorporates the boundary constraint in model evaluation that is necessary to avoid the blind trust in automated PAs.

Experiments show that at the turn of each evolution, new version PA generated by the highest $RIC$ model is always superior to that by the highest $mIoU_{PA}$ model. Also, we stop the iterative PA evolution when the highest $RIC$ for each version converges.

\section{$\mathcal{WS}^2$ for Weakly-Supervised VAAS} \label{sec:vaas}

In the typical VOS, each pixel is assigned an object label, whereas in VAAS it is an actor-action label. To adapt the VOS-oriented $\mathcal{WS}^2$ framework to VAAS, we add another branch in Stage-1 for the additional action label as shown in Figure~\ref{fig:meth_framework}. Hence, apart from the actor-guided attention map that is generated in the same way as in weakly-supervised VOS by a 2D-Conv GCAM, a 3D-Conv GCAM is proposed to generate the action-guided attention map. After binary thresholding, we take the union of the actor and action masks, $M_{\textrm{init}}=M_{\textrm{actor}} \bigcup M_{\textrm{action}}$, as the initial mask. Next, following the same steps in Section~\ref{sec:pseudo_gt}, we refine the blob-like mask $M_{\textrm{init}}$ with SLIC~\cite{slic_radhakrishna_pami12} to produce PA.v0.

\begin{figure}
	\begin{center}
		\includegraphics[width=1\linewidth]{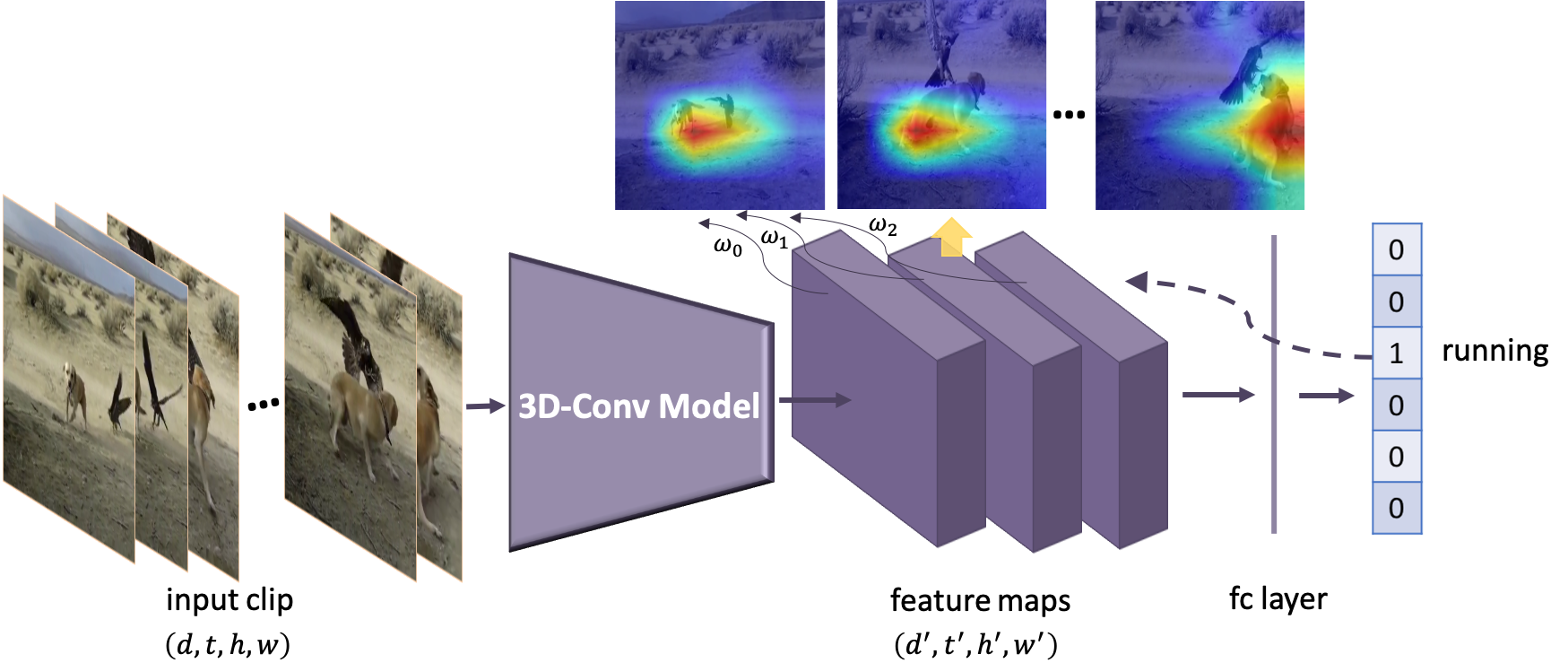}
	\end{center}
	\caption{\textbf{3D-Conv GCAM.} Take \textit{dog-running} as an example, the 3D-Conv Network takes one video clip (consisting of $t$ frames) and the video-level action label, \ie, \textit{running}, as inputs. During back-propagation, the gradients of all classes are set to zeros, except \textit{running} to 1. In total, $t'$ action-guided attention maps corresponding to $t'$ frames uniformly sampled from the input clip are generated to estimate a sparse trajectory of the \textit{running} dog.}
	\label{fig:meth_gcam}
	\vspace{-10pt}
\end{figure}

To implement 3D-Conv GCAM for the given action label, we first obtain a well-trained action classification network denoted as 3D-Conv Model in Figure~\ref{fig:meth_gcam}. Then, we conduct 3D-Conv GCAM to produce action-guided attention maps with the trained models.

\noindent
{\bf Actor-action attention map generation.} \quad  GCAM~\cite{grad_selvaraju_iccv17} is very popular in weakly-supervised learning~\cite{object_wei_cvpr17, weakly_hong_cvpr17, bootstrapping_shen_cvpr18, weakly_qizhu_eccv18}, as it can locate the most appearance-discriminate region purely using the classification network trained with the image-level label. 
We extend GCAM from 2D to 3D to produce the action-guided attention maps for VAAS. 

As shown in Figure~\ref{fig:meth_gcam}, the action attention map is calculated as the weighted average of the feature maps in the last convolutional layer. Specifically, for the target action class $c$, a one-hot vector $y^c$ is back-propagated to the feature maps $\{A_{m}\}$ of the last convolutional layer, the weight $w^c_m$ is the gradient with respect to the $m^{th}$ feature map $A_{m}$:
\begin{equation} \label{eq:weight}
\begin{aligned}
\omega^{c}_{m} = \frac{1}{Z} \sum_{i}{\sum_{j}\sum_{k}{ 
		\frac{\partial y^c}{\partial A_m^{ijk}}}}
\enspace,
\end{aligned}
\end{equation}
where $Z$ is a normalization factor. Once the weights are obtained, the action-guided attention map $\textbf{S}^{c}_{\textrm{action}}$ can be calculated by:
\begin{equation} \label{eq:att_map}
\begin{aligned}
\textbf{S}^{c}_{\textrm{action}} = ReLU(\sum_{m=1}^{d'}{\omega^{c}_{m} A_m})
\enspace.
\end{aligned}
\end{equation}
Compared with 2D-Conv CGAM, each $A_{m}$ is a 3-dimensional feature map of size $(t', h', w')$ with an additional time dimension, thus the obtained $\textbf{S}^{c}_{\textrm{action}}$ is also of size $(t', h', w')$, which can be split into $t'$ attention maps $\{S^{c}_{\textrm{action}}\}^{t'}$. A non-trivial question is how to find the most critical $t'$ (out of $t$) input frames that stimulate the response in the action-guided attention maps. Our empirical findings suggest that a uniform sampling works the best.


\begin{figure}
	\centering
	\includegraphics[width=1\linewidth]{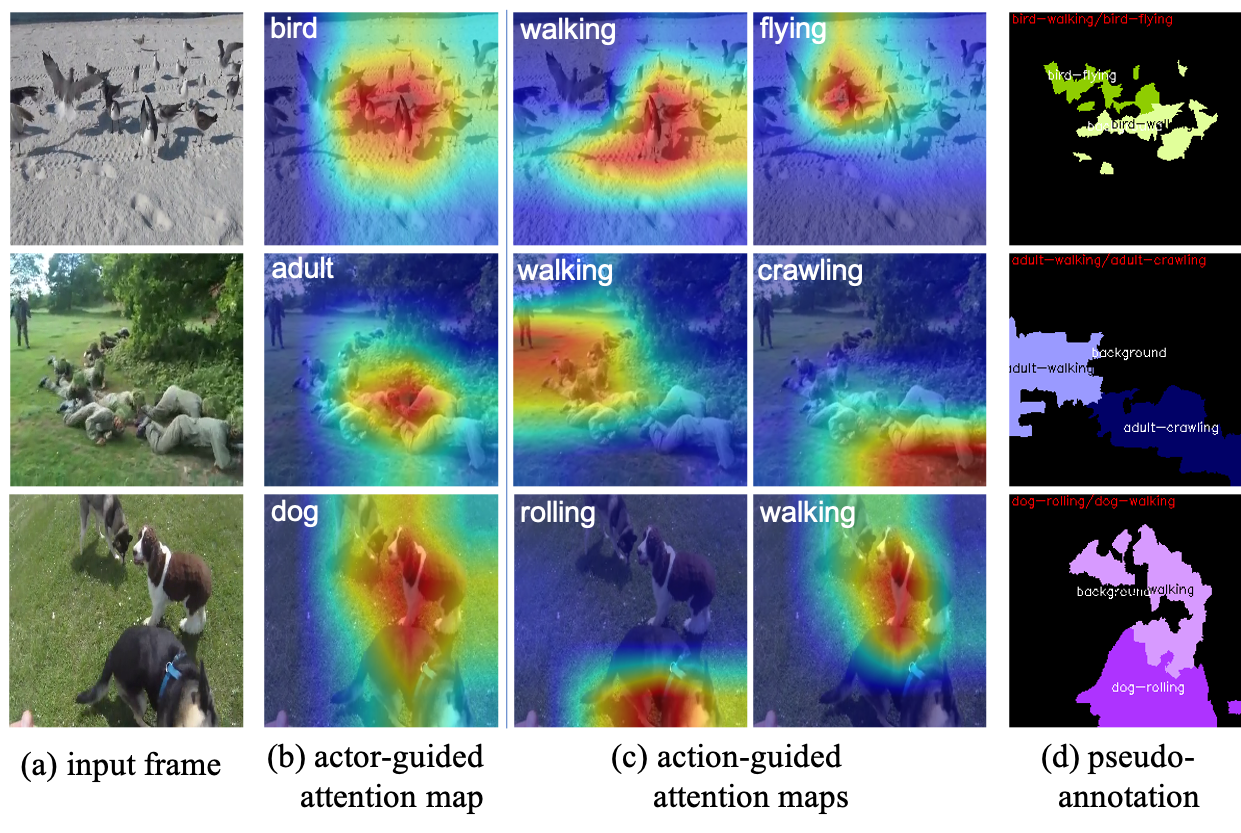}
	\caption{\textbf{Action-guided attention maps help distinguish \textit{single-actor} + \textit{multi-action} cases.}}
	\label{fig:multi_label}
	\vspace{-12pt}
\end{figure} 

\noindent
{\bf Discriminate multiple instances.} \quad One benefit of using 3D-Conv GCAM as the initialization for weakly-supervised VAAS is its ability to distinguish multiple instances. Unlike the \textit{instance} definition in \cite{weakly_zhou_cvpr18, learning_ronghang_cvpr18}, instances here may be the same-type actors doing different actions or vice versa. For some easier scenes that contain multiple different actors, we can set $1$ to the interested actor type in the one-hot vector $y^c$, and localize it only with the actor-guided attention map. However, the actor-guided attention map cannot discriminate actors of the same actor type but performing different actions. As illustrated in Figure~\ref{fig:multi_label}, showing a flock of birds on the beach, some are walking while others are flying. In this case, \textit{walking}-guided and \textit{flying}-guided action attention map will highlight different regions, which enables us to assign the action label to the corresponding actors.

In light of these observations, we further applied 3D-Conv GCAM to weakly-supervised spatial-temporal localization on AVA dataset in Section~\ref{subsec:ava_inter_det}. It turns out that 3D-Conv GCAM shows great potential to focus on the object the person interacts with.


\section{Experiments}

In this section, we first present the quantitative and qualitative performance of the proposed $\mathcal{WS}^2$ on A2D for weakly-supervised VAAS and on YouTube-Object for weakly-supervised VOS. Then we apply the 3D-Conv GCAM to frame-level weakly-supervised action localization on a subset of the AVA dataset to demonstrate its appealing potential in person-object interaction detection.

\subsection{Datasets}
\label{subsec:datasets}

\noindent
{\bf A2D}~\cite{can_xu_cvpr15} is an actor-action video segmentation dataset containing 3782 videos. Different from classic video object segmentation datasets~\cite{a_federico_cvpr16, the_sergi_arXiv18, imagenet_olga_ijcv15, supervoxel_suyog_eccv14}, A2D assigns \textit{actor-action} to the mask, \eg, \textit{cat-eating}. In total, 7 actors and 9 actions are involved. The dataset is quite challenging in terms of unconstrained video quality, action ambiguity, and multi-actor/action, etc. We split the 3036 training videos into two parts, 2748 for training and the rest for validation. 

\noindent
{\bf YouTube-Object}~\cite{discriminative_kevin_cvpr13, supervoxel_suyog_eccv14} consists of 5507 video shots that fall into 10 object classes. Among them, 126 video shots that have
pixel-wise ground-truth annotation in every 10th frame~\cite{supervoxel_suyog_eccv14} are used for testing, and the rest are for training following the same common setting in ~\cite{saliency_wenguan_cvpr15, semantic_yu_cvpr15, semantic_yihsuan_eccv16, spftn_dingwen_cvpr17}. 

\noindent
{\bf AVA}~\cite{ava_gu_cvpr18} is densely annotated with bounding boxes locating the performers with actions.
Videos that fall in 10 classes with evident interactions and a balanced amount of training data are selected for weakly-supervised action localization.We denote it as AVA-10 hereafter.\footnote{The selected classes are fight/hit (a person), give/serve (an object) to (a person), ride, answer phone, smoke, eat, read, play musical instrument, drink, and write.}

\subsection{Implementation Details}

In general, weakly-supervised VOS and VAAS share the two-stage framework, except that the latter has an extra action recognition network in initial PA generation of Stage-1 to account for the action label.

\noindent
{\bf Initial PA generation.} \quad
For A2D, the 2D- and 3D-GCAM are implemented with ResNet-50~\cite{deep_kaiming_cvpr16} pretrained on ImageNet~\cite{imagenet_olga_ijcv15} for actor classification, and inflated 3D ConvNet (I3D)~\cite{quo_joao_cvpr17} pretrained on Kinetics-400~\cite{kinetics_will_arxiv17} for action recognition. To finetune the two models on the A2D, 2794 videos with a single-actor label are used in train \& validation set to train a ResNet-50, and 2639 videos with a single-action label are selected to train an I3D. Once they are well-trained---ResNet-50 achieves 87.74\% accuracy on the single-actor test set, and I3D achieves 76.60\% accuracy on the single-action test set---we apply the two classification networks to its respective GCAM settings for actor-/action-guided attention map generation. Next, the binarized attention masks are refined by SLIC with the thresholds set to $\alpha = 0.5$, $\beta = 0.4$.
For YouTube-Object, we follow the similar procedure, except that only ResNet-50 is used for object classification and attention map generation.

\noindent
{\bf Iterative PA evolution.} \quad 
To select a subset of high-quality PAs, a small network with five layers of Conv-LeakyRelu is constructed to discriminate original foreground patches from cut-and-paste patches. Note that the ResNet-50 trained in Stage-1 is directly used here in testing mode to predict the actor type for the cropped patches.

As for segmentation network, we choose DeepLab-v2~\cite{deeplab_liang_pami18}. During training, the inputs to the network are patches of size $224\times224$ pixels randomly cropped from the frame. We use the ``poly'' learning rate policy as suggested by~\cite{deeplab_liang_pami18}, with base learning rate set to $7\times10^{-4}$ and power to 0.9. We fix a minibatch size of 12 frames, momentum 0.9.
In the testing, we output the full-size segmentation map for each frame. A simple action-alignment post-processing is used to unify the action label for the same actor, since frame-based segmentation network can hardly capture the temporal information throughout the video, which may cause action-inconsistency in the same actor appearing in multiple frames. To tackle this issue, we take the poll of neighboring frames, and assign the action label with the maximum votes to the actor of interest. This procedure is similar to the effective temporal segment network~\cite{temporal_limin_eccv16}, which belongs to the video-level action recognition, whereas ours is in instance-level.

\noindent
{\bf Evaluation metrics.} \quad
We use mean intersection over union ($mIoU$) averaged across all classes to evaluate the performance of the model. 
To compare with the weakly-supervised model~\cite{weakly_yan_cvpr17} on A2D, we also adopt average per-class pixel accuracy ($cls\_acc$) and global pixel accuracy ($glo\_acc$) for evaluation.
%

\subsection{Weakly-Supervised VAAS \& VOS} \label{subsec:a2d_seg}
We first investigate the effectiveness of the key components in iterative PA evolution on A2D. Then we compare our $\mathcal{WS}^2$ model with other state-of-the-art fully- and weakly-supervised methods on A2D and YouTube-Object. Results show that $\mathcal{WS}^2$ outperforms all video-level weakly-supervised models with the performance that is highly competitive even against the fully-supervised models.

\begin{table}
	\small
	\centering
	\begin{tabular}{l|c c|c c}
	\toprule
	\multirow{2}{*}{Models}     & \multicolumn{2}{c|}{Settings}                       & \multicolumn{2}{c}{$mIoU_\textrm{GT}$ (actor-action)}                        \\ \cline{2-5} 
	 & train set     & model eval      & val       & test   \\
	 \midrule
	Baseline   &full    &$mIoU_\textrm{PA}$      &24.62      &20.38 \\ 
	Model-S    &subset  &$mIoU_\textrm{PA}$      &27.65      &24.84  \\ 
	$\mathcal{WS}^2$ &subset  &\textit{RIC}       &\textbf{29.32}     &\textbf{26.74}  \\
	\bottomrule
	\end{tabular}
	\vspace{1pt}
	\caption{\textbf{Comparison of model variants with different settings in Stage-2.} The settings specify whether the model is trained on PAs from the \textit{full} training set or only the selected \textit{subset}. And in each PA version upgrade, whether model with the highest validation $mIoU_\textrm{PA}$ or \textit{RIC} is selected to predict the next version of PA.}
	\label{tab:ablation_study}
	\vspace{-10pt}
\end{table}

\subsubsection{Ablation Study}
\label{subsubsec:ablation_study}

The iterative PA evolution is running in \textit{select-train-predict} cycles, in which \textit{train} is no more special than training a segmentation network as in the fully-supervised setting. The two key factors that influence how much PA can be improved iteration by iteration mainly reside in 1) the overall quality of the \textit{selected} training samples compared with the original full set, and 2) the performance of the model chosen by $RIC$ to predict the next version of PA, compared with that chosen by plain $mIoU_\textrm{PA}$. To quantitatively evaluate their respective contribution to the final model, we conduct an ablation study with three model variants in Table~\ref{tab:ablation_study}.

The results show that, Model-S trained on the subset outperforms Baseline trained on the full PAs, because the selected training samples are of higher-quality. It is also verified in Table~\ref{tab:PA_eval_miou} with the training samples evaluated by the real ground-truth. The selected subsets always have higher $mIoU_\textrm{GT}$ than the full set, which means the selected training samples tend to have more clear boundary and complete coverage of the full object. In comparison, there is more noise and inconsistency in the full set, which may confuse the model and impede it from converging. More importantly, models can be trained much more efficiently on the subset than the full set with 65\%-75\% less training frames.

\begin{table}
	\centering
	\small
	\tabcolsep=0.11cm
	\begin{tabular}{c|c|c|c}
	\toprule
	version      &model eval   &\#frames  &$mIoU_\textrm{GT}$ (a.-a./actor/action) \\
	\midrule
	\multirow{2}{*}{PA.v0}     &init-full  &56120 &23.31 / 31.97 / 29.26\\
	&init-select  &8243 &25.67 / 33.62 / 31.14\\
	\midrule
	\multirow{3}{*}{PA.v1}     &$mIoU_\textrm{PA}$-full  &56120  &28.58 / 38.21 / 35.67\\
	&\textit{RIC}-full  &56120   &29.27 / 38.92 / 36.86\\
	&\textit{RIC}-select  &14669   &32.99 / 41.47 / 39.33\\
	\midrule
	\multirow{3}{*}{PA.v2}     &$mIoU_\textrm{PA}$-full  &56120 &31.72 / 41.54 / 39.06\\
	&\textit{RIC}-full  &56120   &32.36 / 42.34 / 39.94 \\
	&\textit{RIC}-select  &12455   &33.35 / 42.27 / 41.16\\
	\midrule
	\multirow{3}{*}{PA.v3}     &$mIoU_\textrm{PA}$-full &56120  &33.05 / 42.84 / 41.07\\
	&\textit{RIC}-full  &56120   &33.64 / 43.99 / 42.22 \\
	&\textit{RIC}-select  &18330   &34.76 / 43.60 / 42.31\\
	\bottomrule
	\end{tabular}
	\vspace{1pt}
	\caption{\textbf{Quantitative comparison of PA on full/selected training samples produced by models with the highest $mIoU_\textrm{PA}$ or the highest \textit{RIC}.} Here, a.-a. denotes actor-action.}
	\label{tab:PA_eval_miou}
	\vspace{-2pt}
\end{table}

Employing $RIC$ rather than the plain $mIoU_\textrm{PA}$ also helps us choose better models in each PA version upgrade. $mIoU_\textrm{PA}$ calculated by noisy PAs is not guaranteed to assess the true performance of the model as in the fully-supervised setting. It is possible that models with high validation $mIoU_\textrm{PA}$ may also produce noisy prediction that matches exactly the noise in PAs \cite{spftn_dingwen_cvpr17}. 
To overcome this problem, we propose $RIC$ that considers both $mIoU_\textrm{PA}$ and RII (Eq.~\ref{eq:ric}). RII measures the shape change ratio in mask before and after the SLIC refinement. Since refinement drags the segmentation boundary closer to the real object's boundary, if there is not much change on the original prediction after refinement (\ie, high RII), then it is likely that the original prediction has already produced edge-preserving masks that approximate the ground-truth segmentation maps. Table~\ref{tab:PA_eval_miou} clearly exhibits that the-highest-$RIC$ model produces better PAs of the next version than the-highest-$mIoU_\textrm{PA}$ model.

\begin{table}
	\centering
	\small
	\tabcolsep=0.11cm
	\begin{tabular}{c|c}
	\toprule
	Models      &$mIoU_\textrm{GT}$ (actor-action/actor/action)\\
	\midrule
	GPM+TSP~\cite{actor_xu_cvpr16}  &$\textrm{19.9}_{53.9\%}$ / $\textrm{33.4}_{50.3\%}$ / $\textrm{32.0}_{69.1\%}$\\
	TSMT~\cite{joint_kalogeiton_iccv17}+GBH  &$\textrm{24.9}_{67.5\%}$ / $\textrm{42.7}_{64.3\%}$ / $\textrm{35.5}_{76.7\%}$\\
	TSMT~\cite{joint_kalogeiton_iccv17}+SM &$\textrm{29.7}_{80.5\%}$ / $\textrm{49.5}_{74.5\%}$ / $\textrm{42.2}_{91.1\%}$\\
	DST-FCN~\cite{learning_zhaofan_transmm18} &$\textrm{33.4}_{90.5\%}$ / $\textrm{47.4}_{71.4\%}$ / $\textrm{45.9}_{99.1\%}$\\
	Gavrilyuk \etal~\cite{actor_kirill_cvpr18} &$\textrm{34.8}_{94.3\%}$ / $\textrm{53.7}_{80.9\%}$ / \textbf{$\textrm{49.4}_{106.7\%}$}\\
	Ji \etal~\cite{end_ji_eccv18} &\textbf{$\textrm{36.9}_{100\ \%}$} / \textbf{$\textrm{66.4}_{100\ \%}$} / $\textrm{46.3}_{100\ \%}$\\
	\midrule
	$\mathcal{WS}^2$(model.v0)    &$\textrm{19.4}_{52.6\%}$ / $\textrm{38.5}_{58.0\%}$ / $\textrm{31.0}_{67.0\%}$\\
	$\mathcal{WS}^2$(model.v1)    &$\textrm{25.0}_{67.8\%}$ / $\textrm{47.3}_{71.2\%}$ / $\textrm{36.4}_{78.6\%}$\\
	$\mathcal{WS}^2$(model.v2)    &$\textrm{26.6}_{72.1\%}$ / $\textrm{49.2}_{74.1\%}$ / $\textrm{38.1}_{82.3\%}$\\
	$\mathcal{WS}^2$(model.v3)    &\textbf{$\textrm{26.7}_{72.4\%}$ / $\textrm{49.2}_{74.1\%}$ / $\textrm{38.7}_{83.6\%}$}\\
	\bottomrule
	\end{tabular}
	\vspace{1pt}
	\caption{\textbf{Comparison with the state-of-the-art fully-supervised models on the A2D test set.} The subscript denotes the performance percentage to the best fully-supervised model~\cite{end_ji_eccv18}.}
	\label{tab:fs-sota}
	\vspace{-2pt}
\end{table}

\begin{table}
	\centering
	\small
	\tabcolsep=0.11cm
	\begin{tabular}{c|c|c}
	\toprule
	Models     &\textit{$cls\_acc$}     &\textit{$glo\_acc$} \\
	\midrule
	Yan \etal~\cite{weakly_yan_cvpr17} & 41.7 / - / -  & 81.7 / 83.1 / 83.8                  \\
	Ours (model.v3) &\textbf{43.06 / 49.16 / 35.12} &\textbf{87.10 / 91.30 / 87.44} \\
	\bottomrule
	\end{tabular}
	\vspace{1pt}
	\caption{\textbf{Comparison with the state-of-the-art weakly-supervised model on the A2D test set.} \textit{$cls\_acc$} and \textit{$glo\_acc$} are shown in order of actor-action/actor/action.}
	\label{tab:ws-sota}
	\vspace{-2pt}
\end{table}

\subsubsection{Comparison with the State-of-the-Art Methods}
\label{subsubsec:soa}

{\bf A2D.} \quad We compare our weakly-supervised model with the state-of-the-art fully- and weakly-supervised models on the VAAS task. Table~\ref{tab:fs-sota} indicates that our model evolves iteration by iteration, and eventually achieves about 72\% performance of the best fully-supervised model~\cite{end_ji_eccv18}, which is actually a two-stream method that makes use of optical flow for action recognition, whereas our model only takes RGB frames as input. To make a fair comparison with the only existing weakly-supervised model we know of on A2D, we report in Table~\ref{tab:ws-sota} with the evaluation metric used in ~\cite{weakly_yan_cvpr17}. 

\begin{figure}
	\centering
	\includegraphics[width=1.0\linewidth]{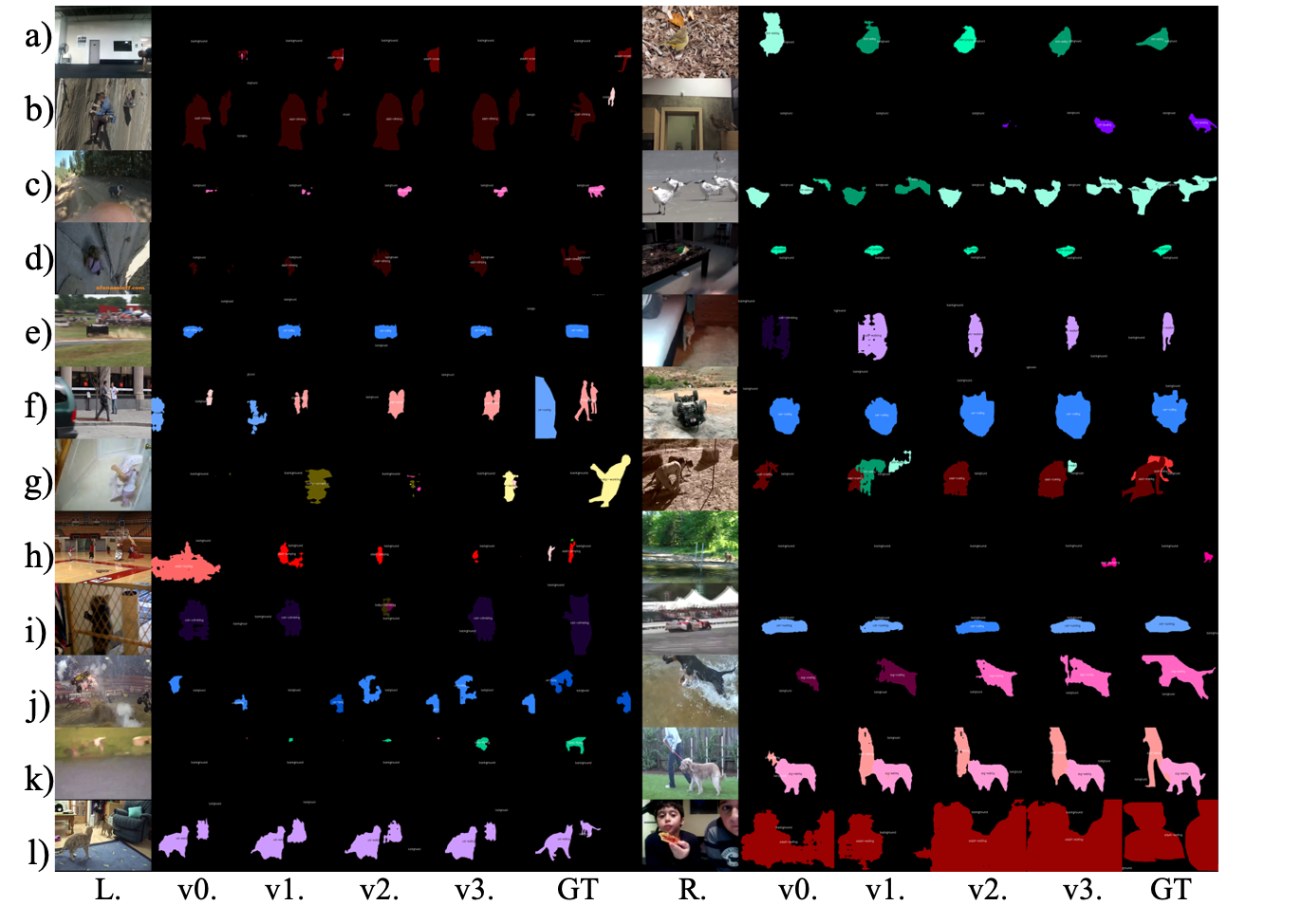}
	\caption{\textbf{Evolution of the model prediction on some tough test samples.} Two samples are shown on each row (left$\rightarrow$right): input frame, prediction by models from v0 to v3, ground-truth (GT). Although in the complete absence of pixel-wise annotation from the GT, our model can still handle challenge cases like occlusion (g-L, i-L, j-L), out of view (a-L, j-R), low illumination ( c-L, b-R, e-R), small objects (d-R, h-R), blur (k-L), multitype-actors (f-L, k-R), fast motion (j-R), background clutter (a-R, g-R), etc.}
	\label{fig:test_pred_evol}
	\vspace{2pt}
\end{figure}

Figure~\ref{fig:test_pred_evol} shows how the model's prediction power evolves through versions. Especially for challenging cases like when only the adult's upper body is observed in a-L, the model correctly predicts \textit{adult-crawling} by seeing his arms erected on the floor. The case of a boy covering his head with a towel (g-L) really gives our model a hard time in the beginning, and it finally figures it out in model.v3. In other hard cases, such as background clutter, motion blur, low illumination, occlusion, out of view, and small scale, the model sometimes fails to output something reasonable in its early versions. Its ability gradually grows as the PA evolves, and finally it gets the prediction right. As for the less complicated cases, the model is able to catch the approximate location of the actors in its early versions, but the predicted masks may suffer from under-/over-segmentation, or wrong action label, which are self-corrected by later versions (see more examples in the supplementary material).

\begin{table}
	\centering
	\footnotesize
	\tabcolsep=0.07cm
	\begin{tabular}{c|c c c c c c c c|c}
	\toprule
	Models 
	&\cite{discriminative_kevin_cvpr13}
	&\cite{video_dong_cvpr13}
	&\cite{fast_anestis_iccv13}
	&\cite{saliency_wenguan_cvpr15} 
	&\cite{semantic_yu_cvpr15}
	&\cite{semi_huiling_accv16}
	&\cite{semantic_yihsuan_eccv16}
	&\cite{spftn_dingwen_cvpr17}
	&$\mathcal{WS}^2$ \\
	\midrule
	$mIoU$ &23.9 &39.1 &46.8 &47.7 &54.1 &60.4 &62.3 &63.1 &\textbf{64.7}\\
	\bottomrule
	\end{tabular}
	\vspace{1pt}
	\caption{\textbf{Comparison with the state-of-the-art video-level weakly-supervised models on the YouTube-Object dataset.}}
	\label{tab:ws-sota-ytb-obj}
	\vspace{-15pt}
\end{table}

\noindent
{\bf YouTube-Object.} \quad $\mathcal{WS}^2$ achieves promising segmentation results which outperform the previous video-level weakly-supervised methods as shown in Table~\ref{tab:ws-sota-ytb-obj}. Qualitative results are given in the supplementary video.

\subsection{Weakly-Supervised Action Localization}
\label{subsec:ava_inter_det}


To further validate the ability of the proposed 3D-Conv GCAM in localizing motion-discriminate part in video, we apply it to weakly-supervised spatial-temporal action localization on  AVA-10. We train an I3D~\cite{quo_joao_cvpr17} action classification network on AVA-10 with only frame-level supervision ({\it without} bounding-boxes). In the testing mode, I3D predicts an action label, with which 3D-Conv GCAM attends to the most relevant region. The visualization of the attention maps \textbf{in the supplementary material} indicates that 3D-Conv GCAM can accurately localize the object the person is interacting with, such as the cigar of a smoking person, or the hand of the person giving/serving something. 

\section{Conclusion}

Given only video-level categorical labels, we tackle the weakly-supervised VOS and VAAS problems. A two-stage framework called $\mathcal{WS}^2$ is proposed to overcome common challenges faced by many synthesize-refine scheme-based methods that are most successful in weakly-supervised VOS. Our proposed select-train-predict cycle utilizes a different cut-and-past model than~\cite{learning_remez_eccv18} to effectively select high-quality PAs and is customized to handle videos. The new region integrity criterion (RIC) is proposed to better guide the convergence of training in the absence of ground-truth segmentation. 
Extensive experiments on A2D and YouTube-Object show that $\mathcal{WS}^2$ performs the best among weakly-supervised methods. Our proposed framework and techniques are general and can be used for other weakly-supervised video segmentation problems. 

\noindent \textbf{Acknowledgments.} This work was supported in part by NSF 1741472, 1813709, and 1909912.
The article solely reflects the opinions and conclusions of its authors but not the funding agents.

{\small
\bibliographystyle{ieee_fullname}
\bibliography{egbib}
}

\end{document}